 \documentclass[final,5p,times,twocolumn]{elsarticle}

\usepackage{amssymb}
\usepackage{amsmath}
\usepackage[colorlinks=true,linkcolor=blue,citecolor=blue,urlcolor=blue]{hyperref}
\usepackage{tabularx}      
\usepackage{array}         
\usepackage{booktabs}      
\usepackage{caption}       
\usepackage{multirow}      
\usepackage{float}

\journal{Expert Systems with Applications}

\begin{document}

\begin{frontmatter}


\title{From Pixels to Posts: Retrieval-Augmented Fashion Captioning and Hashtag Generation}

\author[a]{Moazzam Umer Gondal\corref{equal}}\ead{moazzamumar22@gmail.com}
\author[a]{Hamad Ul Qudous\corref{equal}}\ead{hamad.ulqudous@nu.edu.pk}
\author[a]{Daniya Siddiqui}\ead{daniyasiddiqui120@gmail.com}
\author[a]{Asma Ahmad Farhan}\ead{asma.ahmad@nu.edu.pk}

\cortext[equal]{These authors contributed equally to this work.}

\affiliation[a]{
  organization={School of Computing, National University of Computer \& Emerging Sciences (FAST)},
  city={Lahore},
  postcode={54000},
  country={Pakistan}
}

\begin{abstract}
This paper introduces the retrieval-augmented framework for automatic fashion caption and hashtag generation, combining multi-garment detection, attribute reasoning, and Large Language Model (LLM) prompting. The system aims to produce visually grounded, descriptive, and stylistically interesting text for fashion imagery, overcoming the limitations of end-to-end captioners that have problems with attribute fidelity and domain generalization. The pipeline combines a YOLO-based detector for multi-garment localization, k-means clustering for dominant color extraction, and a CLIP-FAISS retrieval module for fabric and gender attribute inference based on a structured product index. These attributes, together with retrieved style examples, create a factual evidence pack that is used to guide an LLM to generate human-like captions and contextually rich hashtags. A fine-tuned BLIP model is used as a supervised baseline model for comparison. Experimental results show that the YOLO detector is able to obtain a mean Average Precision (mAP@0.5) of 0.71 for nine categories of garments. The RAG-LLM pipeline generates expressive attribute-aligned captions and achieves mean attribute coverage of 0.80 with full coverage at the 50\% threshold in hashtag generation, whereas BLIP gives higher lexical overlap and lower generalization. The retrieval-augmented approach exhibits better factual grounding, less hallucination, and great potential for scalable deployment in various clothing domains. These results demonstrate the use of retrieval-augmented generation as an effective and interpretable paradigm for automated and visually grounded fashion content generation.

\end{abstract}




\end{frontmatter}


\section{Introduction}

Fashion images on social media frequently depict multiple garments styled together and captured in unconstrained conditions, where composition, lighting, and pose are optimized for human appeal rather than machine perception. The central problem is to automatically produce a caption and a set of hashtags that are faithful to the visual evidence: the text must name the detected garment categories, correctly reflect salient attributes such as dominant colors and likely fabric, and adopt language that suits social platforms and brand tone. Despite steady progress in generic image captioning, recent surveys document persistent difficulty with fine-grained attribute grounding and multi-object compositionality, which leads to omissions or hallucinated object–attribute pairings \cite{thakare2024survey,xu2023deep}. For example, when two similarly textured garments co-occur, captioners may collapse them into a single mention or attach the wrong color term to the wrong item; when accessories are partially occluded, they may be ignored or spuriously introduced \cite{jia2023scenegraphs}. In creator and brand workflows, such errors translate into captions that read well but fail to reflect what is actually in the photo, and into hashtags that skew generic rather than representative of color, material, or audience \cite{gusain2023instacaption,jafarisadr2023tags}.

The significance of this issue is that captions and hashtags serve as light-weight, consumer-facing metadata and have a direct impact on search, recommendations, and audience targeting in fashion ecosystems. Offering industry-facing analyses, it is claimed that social streams are one of the most important leading indicators of trend dynamics and merchandising signals, and this has a definite implication on planning and content strategy \cite{balloni2024social4fashion}. Tag suggestion studies show that context-aware and structured tags enhance discoverability and engagement through matching user intent with platform discovery and ranking of tags \cite{jafarisadr2023tags}. A similar role is played by retrieval throughout the fashion pipelines: product discovery relies on visual similarity search, near-duplicate curation, and attribute propagation, and retrieval offers a natural interface between perception and language supervision when explicit labels are sparse or noisy \cite{islam2024fashionretrieval}. In practice, the brands have to work both in the studio catalog imagery and in-the-wild social posts, a deployable solution needs to have strong perception of a variety of garments and local colors, and controlled generation that can give short and style-consistent copy and balanced mix of broad and niche tags, at latencies that fit within real-time or near-real-time content processes.

Literature shows a trajectory from CNN–RNN encoder–decoders toward transformer-based vision–language models. Surveys catalogue advances in attention mechanisms, large-scale pretraining, and decoding strategies, while noting weaknesses in attribute specificity and compositional generalization that surface in complex scenes \cite{thakare2024survey,xu2023deep}. Methods that enrich the visual representation with relational structure (e.g., scene graphs) mitigate certain error modes by making object–relation patterns explicit; nonetheless, they often struggle to bind the correct attribute tokens to the correct instances when similar items co-occur or partially occlude one another \cite{jia2023scenegraphs}. In fashion-specific settings, works that explicitly align detected visual attributes with language—conditioning generation on item categories and attributes—report improved faithfulness and reduced drift from the image content, suggesting attribute grounding is central to domain realism \cite{tang2023fashionattr}. Complementary research treats captioning as a retrieval-informed task, retrieving semantically close exemplars or summaries to stabilize wording, reduce hallucination, and better reflect the target register and platform constraints \cite{mahalakshmi2022summarization}. On the social side, studies of caption and hashtag recommendation reinforce the value of controlled phrasing and structured tags for visibility, intent-signaling, and downstream analytics \cite{gusain2023instacaption,jafarisadr2023tags}. Together, these threads motivate designs that unify multi-object perception, explicit attribute extraction, and retrieval-informed generation, rather than relying on a single monolithic captioner.

Addressing this problem in practice introduces several challenges that shape the design space. First, multi-garment perception is intrinsically difficult: garments overlap, self-occlude, and deform, and recall must be preserved without admitting spurious detections that pollute downstream text. Analyses of modern one-stage detectors highlight sensitivity to confidence thresholds, non-maximum suppression, and class imbalance, underscoring a precision–recall trade-off that must be carefully managed in fashion scenes \cite{lavanya2024yolo,kang2023yolo}. Second, attribute fidelity depends on localized evidence: dominant colors should be computed on garment crops rather than the whole image to avoid background bias, and fine-grained cues such as likely fabric or audience frequently require auxiliary signals beyond the detector’s category label. Fine-grained garment segmentation and fashion classification studies emphasize the value of instance- or part-level delineation to recover attributes that are visually subtle but semantically important \cite{he2023fashionmaskrcnn,chang2022styleyolo}. Third, domain shift between studio catalog photos and social posts degrades out-of-domain performance; cross-domain evaluations regularly report substantial drops without explicit adaptation or retrieval-based conditioning \cite{sirisha2022acrossdomains}. Finally, deployment requires robustness testing and diagnostics to surface failure modes such as attribute hallucination, missed detections, or brittle decoding behaviors, so that thresholds and safeguards can be tuned for real settings \cite{yu2022metaic}.

We address these challenges with a retrieval-augmented formulation that modularizes perception and generation, passing structured evidence between them. Our approach integrates multi-object detection to identify garment instances, localized color estimation to anchor color words, and retrieval against a curated catalog to supply stable attribute cues and stylistic examples. This modular framework allows the system to generate fluent captions and diverse hashtags by conditioning on a compact evidence pack that enumerates detected categories, localized colors, and retrieved attributes. Importantly, the retrieval step serves as a stability prior, reducing attribute hallucination and ensuring diversity in linguistic output without requiring the language generator to be retrained. 

To the best of our knowledge, this work is the first to focus on South Asian fashion using a pipeline that combines multi-garment detection, color extraction, and retrieval-augmented captioning. We present a unique dataset and framework tailored to address the distinctive attributes of South Asian apparel, an area that has been underrepresented in contemporary vision–language research and training corpora. 

We make the following contributions within the fashion captioning domain:
\begin{itemize}
    \item A unified, deployment-oriented pipeline that integrates multi-garment perception, localized color extraction, and retrieval-informed language generation for fashion social media, designed to reduce attribute hallucination and improve the coverage of color, fabric, and target audience.
    \item A new annotated dataset of over 2{,}500 South Asian apparel images, curated for multi-garment detection and captioning research, addressing a region and style underrepresented in current vision–language datasets.
    \item A retrieval module that consolidates weak attributes via similarity-weighted voting and supplies near-neighbor style hints, improving robustness across domain shifts between catalog imagery and social media content.
    \item An empirical comparison with a supervised captioning baseline trained on our dataset, highlighting the strengths and limitations of retrieval-augmented generation in the context of a fashion domain that remains underexplored in the literature, accompanied by robustness diagnostics inspired by modern testing frameworks.
\end{itemize}

\section{Related Work}


Image captioning has evolved into a central task in vision–language research, aiming to translate visual information into coherent natural language. Over the past decade, progress in deep learning has transformed caption generation from template-based and retrieval-driven methods into data-driven architectures that learn cross-modal alignments between vision and text. Early frameworks relied on convolutional and recurrent neural networks to extract image features and sequentially decode captions, while recent transformer-based models leverage self-attention to capture richer global context and long-range dependencies. Surveys of the field describe this evolution and highlight how attention mechanisms, pre-trained vision–language encoders, and multimodal embeddings have significantly improved fluency and contextual relevance in generated captions \cite{thakare2024survey,xu2023deep}. Despite these advances, challenges persist in fine-grained attribute grounding, object–relation reasoning, and domain adaptation, particularly in specialized domains such as fashion or scientific imagery. The broader literature now integrates hybrid pipelines, retrieval augmentation, and domain-specific modeling to address these gaps, reflecting a growing effort to bridge perception accuracy with linguistic expressiveness in modern captioning systems.


The evolution of image captioning has progressed from early encoder–decoder architectures toward transformer-based multimodal frameworks. Classical models coupled convolutional neural networks with recurrent decoders to map image features into textual sequences, achieving foundational success on datasets such as MS COCO and Visual Genome \cite{thakare2024survey,ghandi2023deep}. Attention mechanisms were later integrated to dynamically focus on relevant visual regions during word generation, substantially improving alignment between localized objects and corresponding linguistic tokens \cite{xu2023deep}. These developments established the core paradigm for automatic caption generation and led to measurable gains across standard evaluation metrics including BLEU, METEOR, and CIDEr \cite{thakare2024survey}. Recent transformer-based architectures further advanced this field by modeling global dependencies through self-attention, enabling richer context understanding and more coherent sentence generation across diverse domains \cite{ghandi2023deep,xu2023deep}.

Beyond architectural evolution, several frameworks sought to enhance semantic granularity by decomposing visual scenes into localized entities. Dense-CaptionNet introduced a region-based captioning pipeline that first generated object-level and attribute-level descriptions before merging them into comprehensive sentences, resulting in more accurate portrayals of complex or cluttered images \cite{khurram2021densecaps}. The design allowed simultaneous modeling of multiple regions and relationships, outperforming earlier holistic captioners on datasets such as Visual Genome and MS COCO. Similarly, multi-network ensembles that combine multiple convolutional encoders with natural language modules have been proposed to reduce the semantic gap between visual perception and sentence generation. By fusing diverse feature maps prior to decoding, these systems generate richer and more detailed captions, demonstrating improved descriptive precision over single-model approaches \cite{rinaldi2023multinetwork}.

Despite these advances, surveys consistently report persistent issues such as object hallucination, missing attributes, and limited transferability to unseen domains \cite{thakare2024survey,ghandi2023deep,xu2023deep}. Furthermore, conventional n-gram metrics often fail to capture semantic fidelity and human judgment of caption quality, motivating the exploration of alternative evaluation measures and more structured visual–language alignment strategies \cite{xu2023deep}. Collectively, these studies highlight both the achievements and limitations of generic captioning systems and establish the technical foundation for recent domain-specific and retrieval-augmented approaches that aim to generate attribute-grounded and contextually faithful captions.


An expanding literature is modifying image captioning models to specific fields where visual semantics and vocabulary are not the same as in everyday images. These domain-based systems alter encoders, decoders or datasets to obtain fine-grained details as well as professional language. In fashion, the ability to incorporate structured attribute data in captioning has shown a great deal of success in enhancing descriptive and stylistic relevance. An attribute alignment module was presented in an attribute study, and this links grid-level image features to annotated clothing attributes and a fashion language model trained on a balanced corpus to maintain less vocabulary bias. This combination generated a variety of attribute-based captions and did better with fashion datasets FACAD and Fashion-Gen, proving that the generation of text based on visual attributes improves fidelity and contextual richness \cite{tang2023fashionattr}.

The similar approach can be found in other specialized areas that demand technical accuracy. A hybrid model of convolutional feature extraction with word embeddings trained on geoscience literature is used in geological image captioning, where the decoder is encouraged to use mineralogical terminology through the choice of words. Domain-specific embeddings enhanced semantic accuracy and linguistic suitability compared to generic baselines through integration with domain-specific embeddings, as well as both semantic and linguistic accuracy and appropriateness scores \cite{nursikuwagus2022geocap}. An adaptive attention model in civil engineering that was modified to bridge inspection images focused on areas with damage, e.g., cracks or corrosion, and generated captions that were similar to engineer evaluations both in terminology and spatial resolution to the damage area \cite{li2024bridgecaption}. These methods indicate that the correspondence of attention mechanisms and vocabularies to domain cues generates descriptions that satisfy experts.

Other studies generalize captioning to situations that need general spatial reasoning. A remote sensing pipeline that used U-Net to segment and then caption a scene enhanced the interpretation of the scene, concentrating attention on meaningful land-use categories, resulting in more informative and understandable descriptions of aerial imagery \cite{elsady2023remotesensing}. A comparison of Vision Transformer and VGG encoders with satellite captioning revealed that ViT global self-attention enhances contextual consistency and boosts BLEU and CIDEr scores compared to CNN-based features \cite{han2024vitvgg}. Cross-domain tests also demonstrate the sensitivity of architecture: a comparative study of fashion, art, and medical data indicated that LSTM-based decoders are more likely to be consistent in generalizing to heterogeneous types of images than purely Transformer-based decoders do when adapting to heterogeneous data intrinsically \cite{sirisha2022acrossdomains}. Taken together, these works confirm that the domain adaptation, attribute grounding, and adaptive attention mechanisms significantly enhance caption quality, which provides a theoretical basis of creating retrieval-enhanced fashion captioning systems, which have to acquire localized attributes and stylistic subtlety.


Accurate captioning in the fashion domain depends on reliable visual perception of garments, making detection and segmentation essential. Modern one-stage detectors such as the YOLO family are widely used for their ability to localize multiple objects in real time while maintaining strong accuracy. Studies describe how YOLO reformulates detection as a single regression problem that directly predicts bounding boxes and class probabilities from full images, enabling fast end-to-end inference \cite{lavanya2024yolo}. Comparative analyses across YOLO versions highlight improvements such as stronger backbone networks, anchor-free designs, and refined non-maximum suppression, enhancing precision even under limited data or hardware resources \cite{kang2023yolo}. These advances make YOLO particularly suitable for fashion scenarios involving multiple garments, accessories, or patterns within a single frame.

Building on these foundations, domain-specific detection models have been introduced for fashion imagery. A study using YOLOv5 demonstrated efficient real-time detection of clothing styles—such as plaid, plain, and striped patterns—on modest hardware while maintaining high mean average precision. The results confirmed YOLO’s suitability for fine-grained fashion recognition, showing that single-stage detectors can balance accuracy and speed even on large-scale image collections \cite{chang2022styleyolo}. Complementary work on garment segmentation extends this detection paradigm by delineating item boundaries at pixel level. A modified Mask R-CNN with multi-scale feature fusion and residual modules improved segmentation of overlapping apparel and complex poses, achieving higher accuracy and cleaner boundaries than the baseline model \cite{he2023fashionmaskrcnn}. Similar semantic segmentation enhancements, including edge-aware metrics and architecture adaptation methods, further refine region boundaries and object shapes, enabling more precise extraction of garment masks \cite{he2022segmentation,tereikovskyi2022method}.

Integration of detection and captioning has also been explored outside fashion. A vision-based system for construction imagery combined YOLO-style detection with a caption generator to produce structured scene descriptions, demonstrating that localized object features enhance textual understanding \cite{wang2022construction}. These findings collectively underscore that robust detection and segmentation modules form the perceptual backbone of any captioning pipeline. For fashion captioning, they provide the essential groundwork for isolating multiple garments, capturing localized color information, and ensuring that language generation aligns with the visual evidence present in each region.


The current studies are showing a tendency to incorporate retrieval and hybrid learning methods in the process of image captioning in order to enhance semantic consistency and grounding. These methods are integrated visual recognition, text summary, and optimization to bridge the gap between perception and language generation. A representative research incorporated both summarization and captioning, which involved the integration of BiLSTM-based text encoding and Deep Belief Network to summarize and produce image descriptions concurrently. This multimodal scheme boosts visual context in retrieval outcomes, and the accuracy, recall, and F-scores of summarization and BLEU scores are not far below the human ones, proving that cross-modal fusion increases semantic richness in retrieval-oriented tasks \cite{mahalakshmi2022summarization}.

Models based on optimization expand on this combination by using metaheuristic algorithms to optimize network parameters or the output of the generated networks. A hybrid captioning framework used genetic or particle swarm optimization to optimize deep encoder-decoder models and optimize captions most likely to be descriptive, providing quantifiable and statistically significant improvements in BLEU and CIDEr in comparison to baseline networks. The model was able to generate captions that were syntactically consistent and semantically elaborate by escaping local optima through metaheuristics application and considering the model to be semantically detailed and syntactically well-formed in captioning \cite{alduhayyim2022metaic}. Likewise, the ensemble-based and item-level hybrid methods focus on the integration of specialized feature extractors and attribute reasoning to empower caption semantics. A style classification network based on item region, such as, made use of domain-specific pooling and dual backbones to locate garment area and combine their characteristics using gating processes, enhancing classification precision by up to 17\% compared to baselines \cite{choi2024irsn}. Such advances of item-level feature representation guide captioning architecture, which needs to formulate several visual objects with exact associations.

The retrieval as the concept has become one of the central elements of fashion and visual understanding mechanisms. Surveys on fashion image retrieval classify the current systems as cross-domain, attribute-based, or outline-level retrieval pipelines and refer to the fact that visual similarity embeddings allow their application in product matching and complimentary item recommendation applications \cite{islam2024fashionretrieval}. Supplementary structures combine retrieval and generative language frameworks in the individual styling. A generative AI-based recommendation system was used to generate textual advice about the outfits by hybridizing YOLOv8 detection with the GPT-4 model and achieved good scores in evaluation and user satisfaction in fashion recommendation tasks with localized clothing crops \cite{kalinin2024stylegen}. Combined with these hybrid and retrieval-augmented studies, one can conclude that structured search, optimization, and attribute-level reasoning can generate more informative and context-consistent captions, and is a conceptual basis of retrieval-augmented generation pipelines.


Parallel to advances in vision–language modeling, research has explored caption and hashtag generation for social media, where engagement and contextual relevance are central objectives. These studies combine computer vision, natural language processing, and trend analytics to generate audience-aware textual content aligned with visual cues. A deep learning caption recommendation engine trained on Instagram data demonstrated that integrating visual analysis with neural language generation produces more contextually relevant captions than manual authoring. The system used a convolutional network for image understanding and a language model tuned to social media phrasing, capturing stylistic tone typical of platform communication \cite{gusain2023instacaption}. Complementary work on hashtag prediction proposed a two-stage framework: a ResNet-based classifier detected semantic concepts in images, and a transformer generator produced trending hashtags. This multimodal pipeline increased relevant tag coverage by roughly 11\% compared with baseline approaches, confirming that coupling visual and linguistic representations enhances content visibility \cite{jafarisadr2023tags}.  

Social analytics frameworks extend these ideas to large-scale trend discovery. A fashion intelligence system applied object detection to social media posts, focusing on handbags as a case study, and extracted features such as type and dominant color to identify emerging style patterns. Achieving 97\% classification accuracy and 0.77 mean average precision, it demonstrated how automated analysis of social imagery can inform design and marketing decisions \cite{balloni2024social4fashion}. Beyond caption and tag generation, multimodal discourse modeling examines how images and text interact semantically. A study of cross-modality discourse classified five types of image–text relations—from direct description to conceptual extension—using a multi-head attention model trained on annotated tweet pairs, achieving state-of-the-art accuracy \cite{xu2023crossmodal}. Collectively, these approaches illustrate that blending image understanding with language modeling enhances the creativity, interpretability, and analytic value of social media content, reinforcing the need for captioning systems that balance visual precision with communicative engagement.


Despite extensive progress across captioning, detection, and retrieval research, several limitations remain that constrain real-world deployment in fashion applications. Most existing captioning models, including transformer-based and domain-adapted variants, focus on single-object scenes or rely on pre-defined attributes, limiting their ability to describe multi-garment compositions typical of social media fashion imagery. Detection-oriented studies excel in localizing objects but rarely connect those results to coherent language generation, while retrieval and optimization frameworks often address semantic alignment in isolation rather than integrating visual perception with generative modeling. Current domain-specific works improve attribute grounding but are confined to narrow datasets and lack mechanisms to generalize stylistically across domains. Furthermore, social-media captioning systems prioritize engagement or tag relevance without ensuring factual correspondence to image content. Collectively, these gaps highlight the need for an integrated pipeline that unifies multi-object detection, localized attribute extraction, and retrieval-informed language generation to produce accurate, context-aware, and stylistically adaptive captions for complex fashion scenes.

The remainder of this paper is organized as follows. Section~\ref{sec:method} outlines the proposed methodology, including data preparation, model design, and retrieval-augmented caption generation. Section~\ref{sec:implementation} describes the implementation setup, experimental configuration, and training details for all components. Section~\ref{sec:results} presents the quantitative and qualitative evaluation of the proposed system, covering detection, captioning, and hashtag generation. Section~\ref{sec:D&F} discusses the key findings, limitations, and directions for future work, while Section~\ref{sec:conclusion} concludes the paper with final remarks and implications.

\begin{table*}[!t]
\caption{Condensed summary of representative literature discussed in Sections 2.1–2.5.}
\label{tab:relatedwork_full}
\setlength{\tabcolsep}{4pt}
\renewcommand{\arraystretch}{1.05}
\footnotesize
\begin{tabularx}{\textwidth}{|>{\centering\arraybackslash}p{0.5cm}|p{2.6cm}|p{2.5cm}|X|}
\hline
{\scriptsize \#} & \textbf{Domain / Task} & \textbf{Core Dataset / Context} & \textbf{Method and Key Highlights} \\
\hline

\cite{thakare2024survey} & Generic captioning (survey) & MS COCO, Visual Genome & Taxonomy of CNN–RNN, attention, transformers; datasets \& metrics; challenges incl.\ hallucination and weak attribute grounding. \\ \hline

\cite{ghandi2023deep} & Captioning (survey) & Multiple benchmarks & Structured review; taxonomy and comparative ranking; highlights bias, misalignment, interpretability. \\ \hline

\cite{khurram2021densecaps} & Region-based captioning & Visual Genome, MS COCO & Dense-CaptionNet: region/object and attribute description fused to full sentence; improved detail on complex scenes. \\ \hline

\cite{rinaldi2023multinetwork} & Captioning (ensemble) & Generic image datasets & Multi-network CNN ensemble + NLP decoder; fuses diverse visual features to reduce semantic gap and boost precision. \\ \hline

\cite{xu2023deep} & Captioning (survey) & Multiple benchmarks & Trends from CNN–RNN to Transformers; attention, scene structure; evaluation–human mismatch and real-time constraints. \\ \hline

\cite{tang2023fashionattr} & Fashion captioning & Fashion-Gen, FACAD & Attribute Alignment Module + Fashion Language Model (balanced corpus); improves attribute-grounded, diverse captions. \\ \hline

\cite{nursikuwagus2022geocap} & Geological captioning & Domain-specific geology & CNN + domain word embeddings; injects scientific vocabulary; more accurate, context-appropriate terminology. \\ \hline

\cite{li2024bridgecaption} & Civil engineering & Bridge inspection images & Adaptive attention emphasizes damage regions; captions consistent with engineer reports (location/type). \\ \hline

\cite{elsady2023remotesensing} & Remote sensing captioning & Aerial imagery & U-Net segmentation before captioning; focuses on meaningful regions; clearer land-use descriptions. \\ \hline

\cite{han2024vitvgg} & Remote sensing & Satellite datasets & ViT vs VGG encoders; ViT’s global context improves BLEU/CIDEr and descriptive coherence. \\ \hline

\cite{sirisha2022acrossdomains} & Cross-domain captioning & Fashion, art, medical, news & Comparative study; LSTM decoders generalize better than pure Transformer decoders across domains. \\ \hline

\cite{lavanya2024yolo} & Object detection (review) & Real-time detection & YOLO single-pass regression; high-speed multi-object localization and practical optimization insights. \\ \hline

\cite{kang2023yolo} & Detection/segmentation (review) & COCO evaluations & Evolution of YOLO versions; backbone upgrades, anchor-free ideas, NMS tuning; precision under constraints. \\ \hline

\cite{chang2022styleyolo} & Fashion style recognition & Custom (five patterns) & YOLOv5s detects plaid/plain/striped styles; high mAP and FPS on modest hardware (real-time feasibility). \\ \hline

\cite{he2023fashionmaskrcnn} & Garment segmentation & Fashion datasets & Modified Mask R-CNN with multi-scale fusion/residual modules; cleaner boundaries for overlapping apparel. \\ \hline

\cite{he2022segmentation} & Segmentation (edge-aware) & Generic segmentation & Region-edge metric and loss to improve boundary quality; +1\% overall, +4\% on edge-region metrics. \\ \hline

\cite{tereikovskyi2022method} & Segmentation (methodology) & Automotive images & Procedure to select/adapt encoder–decoder by task constraints; $\sim$80\% accuracy with short training. \\ \hline

\cite{wang2022construction} & Detection + captioning & Construction site images & Detector features fed to captioner; structured captions (scene graphs) for queryable site understanding. \\ \hline

\cite{mahalakshmi2022summarization} & Retrieval + captioning & Gigaword, DUC & BiLSTM text encoding + DBN summarization + image captions; better P/R/F for summaries; captions near human BLEU. \\ \hline

\cite{alduhayyim2022metaic} & Hybrid optimization & Standard caption sets & Metaheuristics (GA/PSO) tune encoder–decoder and outputs; BLEU/CIDEr gains via search beyond local optima. \\ \hline

\cite{choi2024irsn} & Fashion style classification & Fashion style sets & Item-region pooling, dual backbones, gated fusion; up to 16–17\% accuracy gains (avg.\ $\sim$9\%). \\ \hline

\cite{islam2024fashionretrieval} & Fashion image retrieval (survey) & Cross-/attribute/outfit retrieval & Taxonomy of FIR; cross-domain matching, attribute-based search, outfit recommendation; multimodal embedding needs. \\ \hline

\cite{kalinin2024stylegen} & Fashion recommendation & User photos; YOLOv8+LLM & YOLO crops + GPT-based advice; competitive user ratings vs other assistants; localized, personalized styling. \\ \hline

\cite{gusain2023instacaption} & Social captioning & Instagram data & CNN + NLP caption engine; platform-aligned tone; more contextually suitable than manual authoring. \\ \hline

\cite{jafarisadr2023tags} & Hashtag generation & Social images & ResNet classifier + Transformer generator; $\sim$11\% increase in relevant/trending tags over baselines. \\ \hline

\cite{balloni2024social4fashion} & Trend mining & Instagram (handbags) & Detects handbags; extracts type/colors; 97\% classification, mAP 0.77; dashboards for trend discovery. \\ \hline

\cite{xu2023crossmodal} & Image–text discourse & 16k tweets (labeled) & Multi-head attention classifier of five image–text relations; state-of-the-art discourse relation accuracy. \\ \hline

\end{tabularx}
\end{table*}

\section{Methodology}\label{sec:method}

\begin{figure}[htbp]
\centering
\includegraphics[width=0.48\textwidth]{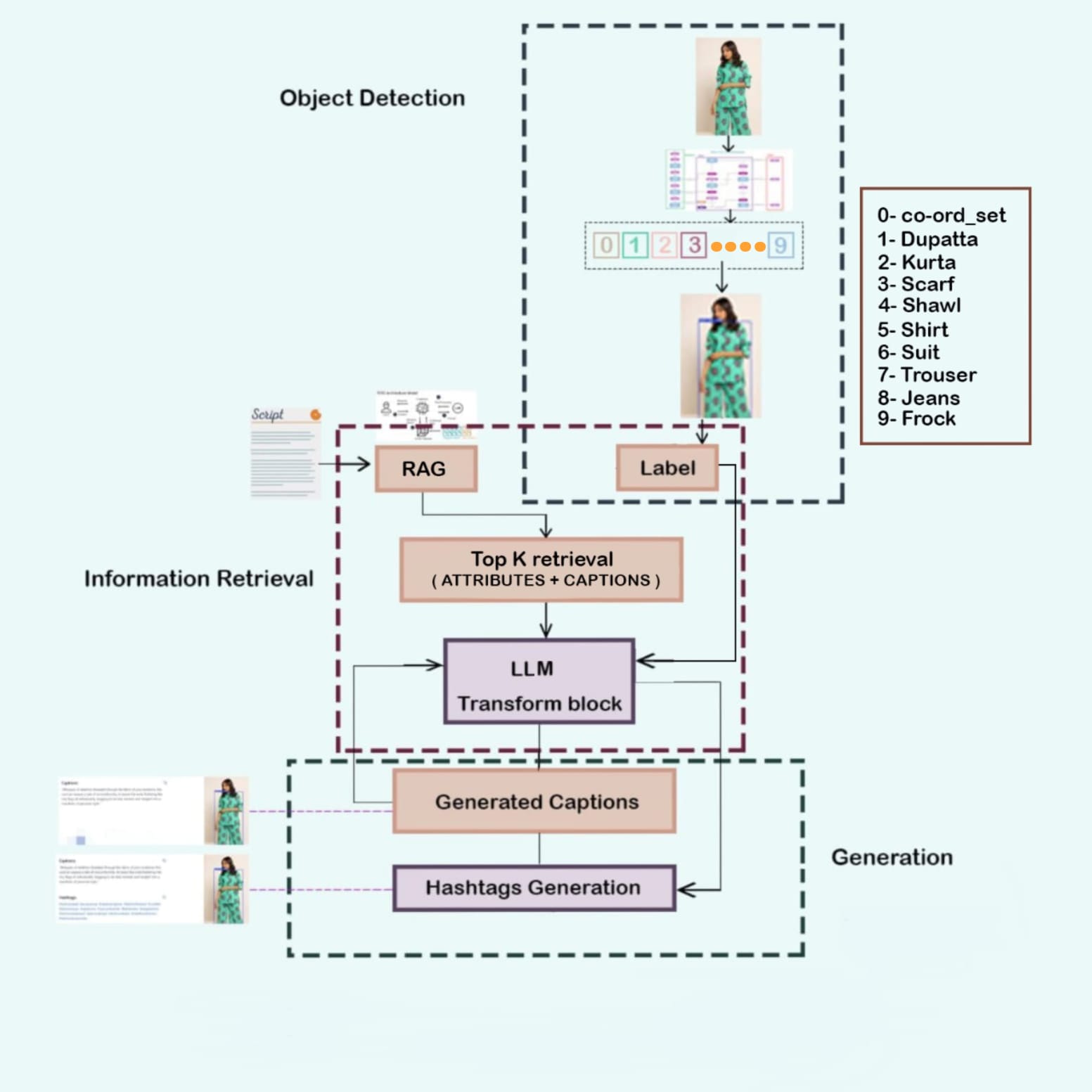}
\caption{High-level architecture of the proposed retrieval-augmented fashion captioning pipeline showing object detection, information retrieval, and generative captioning stages.}
\label{fig:architecture}
\end{figure}

The suggested framework unites the concepts of multi-garment recognition, retrieval-enhanced rationale, and generative caption creation into one fashion image comprehension structure. The system has been structured into three significant stages as demonstrated in Fig.~\ref{fig:architecture}. In the \textit{Object Detection} block, a YOLO-based model detects all garments in the image (e.g., shirt, dupatta, frock, cordset) and labels each of them with a class. The \textit{Information Retrieval} block calculates embeddings of the identified image and retrieves semantically similar examples, integrating the visual evidence by a focused embedding encoding the fabric and gender features. Lastly, the \textit{Generation} block makes use of an LLM to turn the structured evidence into fluent but attribute-based captions and generates a uniform set of hashtags. Together, these modules form a retrieval-augmented pipeline capable of generating visually faithful and stylistically adaptive social media descriptions.

\subsection{Data Preparation and Preprocessing}

The data utilized in the present work was collected in the form of posts in social media and web pages of fashion brands that are publicly available in order to include consumer-style and catalog-quality images. The corpus collected is varied in terms of both category of clothes, poses and lighting conditions, which mirror the diversity of fashion photography in the real world. There is also a subset of pictures that contain metadata, as the titles of a product or a short description, and there are also pictures that contain only raw images. All pictures were downsized to a standard resolution, placed in the RGB color space and filtered to exclude duplicates or low-quality samples. In the case of metadata, text fields were cleansed and tokenized to identify useful properties such as color, gender, and fabrics. The resulting dataset thus integrates multi-source visual data and heterogeneous textual information, forming a balanced foundation for training and evaluating the proposed detection, retrieval, and caption generation modules under varied stylistic and environmental conditions.

\subsection{Object Detection and Visual Attribute Extraction}

The perception stage of the system is responsible for identifying all visible garments in a fashion image. A YOLO-based object detector was fine-tuned on the curated fashion dataset comprising categories such as shirt, dupatta, trouser, frock, and other garments. The model was trained to detect multiple garments per image, enabling multi-label and multi-instance recognition under varied poses and lighting conditions. During inference, all bounding boxes with confidence scores above a fixed threshold $(\theta_{conf})$ and non-maximum suppression IoU threshold $(\theta_{iou})$ are retained to ensure that overlapping garments are accurately localized. Each detected region is then cropped to serve as an independent visual instance for further analysis.

For every cropped detection, the dominant garment colors are estimated using $k$-means clustering in the RGB space. Given pixel samples $\{x_1, x_2, \dots, x_n\}$, the algorithm partitions them into $k$ clusters $\{C_1, C_2, \dots, C_k\}$ by minimizing intra-cluster variance:

\begin{equation}
\underset{C}{\arg\min} \sum_{i=1}^{k}\sum_{x_j \in C_i} \|x_j - \mu_i\|^2,
\end{equation}

where $\mu_i$ represents the centroid of cluster $C_i$. The two largest clusters are chosen as primary and secondary colors after discarding near-white or near-black clusters. Each centroid is mapped to its nearest perceptual color name in the CIELAB color space using Euclidean distance, providing an interpretable and consistent color representation for each garment. The resulting detections and associated color features form the structured visual input for the subsequent retrieval stage.

\subsection{Retrieval-Augmented Attribute Inference}

After the detection and color extraction stages, a retrieval-augmented reasoning module is applied to infer global attributes such as fabric and target gender. The module embeds each image into a joint vision–language space using the CLIP encoder. Given an image $I$, its normalized embedding vector $v$ is computed as

\begin{equation}
v = \frac{f_{\text{CLIP}}(I)}{\|f_{\text{CLIP}}(I)\|_2},
\end{equation}

where $f_{\text{CLIP}}(\cdot)$ denotes the CLIP visual encoder. For each query embedding $v_q$, the similarity to all indexed catalog images $\{v_i\}$ in the FAISS database is obtained through cosine similarity,

\begin{equation}
s(I_q, I_i) = \frac{v_q \cdot v_i}{\|v_q\|\,\|v_i\|}.
\end{equation}

The Top-$K$ most similar items are retrieved and used to perform attribute voting. Let each neighbor $i$ have an associated attribute label $y_i$ (e.g., “cotton”, “female”). A similarity-weighted score for each label $y$ is calculated as

\begin{equation}
\text{score}(y) = \sum_{i:y_i=y} w(s_i), \quad
w(s_i) = e^{\tau s_i},
\end{equation}

where $\tau$ is a temperature parameter controlling the exponential weighting. The final predicted attribute $\hat{y}$ and its confidence $c_{\hat{y}}$ are derived as

\begin{equation}
\hat{y} = \arg\max_y \text{score}(y), \qquad
c_{\hat{y}} = \frac{\text{score}(\hat{y})}{\sum_{y'} \text{score}(y')}.
\end{equation}

If $c_{\hat{y}}$ falls below a defined threshold $(\theta_{attr})$, the attribute is assigned as “unknown.” This retrieval-based voting stabilizes predictions by aggregating evidence from visually similar exemplars rather than relying solely on single-image inference. In addition to attribute inference, the Top-$K$ retrieved samples also provide concise textual snippets—titles or short descriptions—that serve as stylistic cues for the language generation stage.

\subsection{Caption and Hashtag Generation}

The final stage of the pipeline generates descriptive captions and context-aware hashtags by prompting an LLM with structured visual evidence. For each processed image, an evidence pack $E$ is constructed as

\begin{equation}
E = \{D, A, R\},
\end{equation}

where $D$ contains detected garments and their color descriptors, $A$ includes retrieved global attributes such as fabric and target gender, and $R$ represents concise caption examples retrieved from the catalog index. This structured representation is converted into a textual prompt template that guides the model to produce detailed, visually grounded sentences.

A pre-trained language model $\mathcal{F}_{\text{LLM}}(\cdot)$ receives the prompt containing $E$ and generates a fluent caption $\hat{C}$ and a complementary set of hashtags $H$ as:

\begin{equation}
\{\hat{C}, H\} = \mathcal{F}_{\text{LLM}}(E,\,P),
\end{equation}

where $P$ denotes the curated prompt instructions defining tone, format, and length. The caption emphasizes garment attributes and visual harmony, while the hashtag set balances general and specific tags related to color, fabric, and occasion. Since the model operates in a retrieval-augmented mode, linguistic style is guided by the examples in $R$, ensuring descriptive diversity without template repetition.

\subsection{BLIP Baseline for Comparison}

Fine-tuned BLIP model was utilized as a supervised baseline to evaluate the effectiveness of the suggested retrieval-augmented framework. BLIP combines a vision encoder with a language decoder, which are jointly trained on image-text pairs, which allows it to generate captions end-to-end. In this paper, the model was trained on the filtered fashion dataset with the same training-validation splits as the retrieval index. The fine-tuning task minimized the cross-entropy loss between generated and reference captions, enabling the model to acquire brand-specific language patterns and visual semantics. In inference, the BLIP baseline made captions independently of images without extra retrieval or formatted attribute input. This is where the difference between the traditional end-to-end captioners and the suggested modular, retrieval-enhanced model that explicitly adds visual qualities and stylistic examples to caption and hashtag generation lies.

\subsection{Evaluation Metrics and Experimental Design}

In the experimental analysis, the two important parts of the system such as visual perception and text generation were evaluated using standard measures. These tests are measures of the accuracy of the perceptual component in detecting as well as the linguistic quality and diversity of the textual outputs produced.

\subsubsection{Object Detection}  
The YOLO-based detector was evaluated using mean Average Precision (mAP) at IoU thresholds of 0.5 and 0.5–0.95, consistent with the COCO evaluation protocol:

\begin{equation}
\text{mAP} = \frac{1}{N_c} \sum_{i=1}^{N_c} \int_{0}^{1} p_i(r)\,dr,
\end{equation}

where $p_i(r)$ denotes the precision–recall curve for class $i$ and $N_c$ is the number of garment categories. This measure jointly captures localization accuracy and classification precision for multi-garment scenes.

\subsubsection{Caption Quality}  
Retrieval-augmented pipeline and the BLIP baseline were compared in terms of caption fluency and descriptive accuracy measured by BLEU, METEOR, and ROUGE-L. The scores quantify lexical and structural similarity between generated captions and reference descriptions to evaluate the capacity of each model to recreate garment attributes, color terms, and contextual relationships. Besides text measures, a CLIP similarity measure was used to estimate visual-semantic correspondence between images and captions. Using the CLIP~ViT-B/32 encoder, cosine similarity was computed between each image embedding $v_I$ and its caption embedding $v_T$:

\begin{equation}
\text{CLIPSim} = \frac{v_I \cdot v_T}{\|v_I\|\,\|v_T\|}.
\end{equation}

The mean similarity scores were reported for both generated captions and original product descriptions, and their difference $\Delta = \text{CLIPSim}_{\text{pred}} - \text{CLIPSim}_{\text{orig}}$ indicates whether generated captions exhibit stronger or weaker visual correspondence relative to the original text. This combination of lexical and semantic metrics provides a balanced evaluation of linguistic quality and visual grounding across models.

\subsubsection{Hashtag Evaluation}  
For the RAG-LLM system, two complementary metrics were employed to assess the quality of generated hashtags. The first, \textit{attribute coverage}, measures how effectively the predicted hashtags capture key visual facets such as garment category, dominant color, fabric, and target gender. For each image~$i$, an attribute coverage ratio is computed as

\begin{equation}
\text{cov}_i = \frac{\sum_{f \in F_i} \text{hit}_i(f)}{|F_i|},
\end{equation}

where $F_i$ denotes the set of known facets for image~$i$, and $\text{hit}_i(f) = 1$ if any synonym of the facet value appears in the generated hashtags, and $0$ otherwise. Synonym dictionaries are used to normalize linguistic variants such as \textit{men/male/mens} or \textit{woman/women/female}. An image is considered correctly covered if $\text{cov}_i \geq \tau$, where $\tau$ is a threshold (set to $0.5$ in our experiments). The overall coverage is then defined as

\begin{equation}
\text{Coverage@}\tau = \frac{|\{i \mid \text{cov}_i \ge \tau\}|}{\text{Total images}}.
\end{equation}

The second metric, \textit{Distinct-$n$}, quantifies linguistic diversity by computing the ratio of unique $n$-grams to total $n$-grams across all generated hashtags:

\begin{equation}
\text{Distinct-1} = \frac{\text{Unique unigrams}}{\text{Total unigrams}}, \quad
\text{Distinct-2} = \frac{\text{Unique bigrams}}{\text{Total bigrams}}.
\end{equation}

Together, these measures capture both the semantic relevance of hashtags to image content and the lexical diversity of the generated outputs, providing a comprehensive evaluation of social-media-oriented text generation.

\section{Implementation}\label{sec:implementation}

All components of the proposed framework were developed in Python using a modular architecture that integrates perception, retrieval, and language generation within a unified workflow. Experiments were conducted in a cloud-based environment on Google Colab, utilizing an NVIDIA~A100 GPU. The deep learning modules were implemented in PyTorch, with OpenAI CLIP and FAISS used for visual embedding and similarity indexing. The BLIP baseline was implemented using the \texttt{transformers} and \texttt{datasets} libraries from Hugging~Face. LLM prompting was executed via the Groq~API using the LLAMA~3 backbone. The entire pipeline—including YOLO-based multi-garment detection, color extraction, retrieval, BLIP fine-tuning, and caption generation—was executed as modular, reproducible scripts with fixed random seeds and consistent dataset splits across all experimental runs.

\subsection{Dataset Setup and Details}

A total of approximately 3,000 fashion images were collected from various social media pages and official brand websites, covering a wide range of apparel categories, poses, and backgrounds to capture real-world diversity. In addition, a subset of 1,200 catalog-style product images was curated with accompanying metadata containing fields such as product title, textual description, fabric type, and dominant color. The combined dataset thus represented both consumer-generated and catalog imagery, providing a balanced foundation for multi-garment detection and captioning tasks.

All images were manually annotated in Roboflow using bounding boxes and class labels corresponding to primary garment types such as shirt, trouser, dupatta, frock, co-ord set, scarf, suit, shawl and jeans. The annotated data were merged and augmented through Roboflow’s automated pipeline, resulting in 4,725 training images and 1,466 test images. Preprocessing operations included auto-orientation correction and resizing, where all images were stretched to a uniform resolution of 640$\times$640 pixels to standardize the detector input.

Augmentation was applied to increase data diversity and improve model generalization. For each training sample, three augmented outputs were generated using the following transformations: horizontal flips; 90° clockwise and counterclockwise rotations; random crops with 0–20\% zoom; rotations between $-15^{\circ}$ and $+15^{\circ}$; grayscale applied to 15\% of images; brightness variations within $\pm$15\%; Gaussian noise applied to 0.1\% of pixels; and bounding-box perturbations including rotation ($\pm$15°) and blur (up to 2.5\,px). These augmentations produced a visually varied and balanced dataset for robust YOLO training.

\subsection{YOLO Training and Multi-Garment Inference}

A YOLO-based detector was employed to identify multiple garments within each fashion image. The lightweight \texttt{YOLOv11s} model from the Ultralytics framework was initialized with pretrained weights (\texttt{yolo11s.pt}) and fine-tuned on the curated dataset described in Section~4.1. Training was performed using the official \texttt{Ultralytics} package in Python. The network was trained for 100~epochs with an input resolution of 640$\times$640\,px, optimizing both localization and classification heads for all annotated apparel categories. Roboflow-generated augmentation ensured robustness to viewpoint variation, brightness shifts, and partial occlusion. Model performance was monitored through validation mean Average Precision (mAP) at IoU thresholds of 0.5 and 0.5–0.95, selecting the best checkpoint for downstream inference.

During inference, detections were filtered using a confidence threshold of 0.35 and a non-maximum suppression IoU threshold of 0.6 to preserve overlapping garments. All bounding boxes above these thresholds were retained and cropped to produce individual garment regions. These localized crops, together with class predictions, served as inputs for subsequent color extraction and retrieval-augmented attribute inference.

\subsection{Color Extraction and Attribute Processing}

For each detected garment region, localized color analysis was performed to extract the dominant visual attributes that guide retrieval and caption generation. The cropped detections were processed through a $k$-means clustering algorithm implemented with \texttt{scikit-learn}, where the number of clusters was fixed to $k=4$. Pixel values were sampled uniformly within each crop, and clustering was executed in the RGB color space. The two largest clusters, representing the most visually significant colors, were retained as the primary and secondary tones. To suppress noise, near-white and near-black clusters with coverage below 6\% of the region were discarded. Each remaining cluster centroid was converted from RGB to CIELAB coordinates and matched to the closest perceptual color name in a predefined palette using Euclidean distance. The resulting color tokens were appended to the corresponding garment detections, forming structured descriptors used as part of the retrieval-augmented reasoning stage.

\subsection{CLIP Embeddings and FAISS Index Construction}

To support retrieval-augmented reasoning, a structured catalog subset containing 1,195 product images with complete metadata—titles, descriptions, fabric type, color, and gender—was used to construct the retrieval index. The dataset was divided into 80\% training and 20\% test partitions in a category-aware manner, ensuring that each garment class contributed at least one image to both splits. Each image was converted to RGB format and encoded using the CLIP~ViT-B/32 visual backbone. The normalized embeddings were indexed using FAISS, enabling efficient inner-product similarity search equivalent to cosine similarity on normalized vectors. Metadata for each indexed image, including attribute labels and short text descriptions, was stored in accompanying JSONL files for structured access. During inference, the system queried the FAISS index to retrieve the Top-$K$ ($K=20$) most similar items, applying similarity-weighted voting to infer global attributes (fabric and gender) and sampling retrieved textual snippets as stylistic cues for caption generation.

\subsection{LLM Integration for Caption and Hashtag Generation}

The last generation phase was introduced by a FastAPI inference service that would be connected to the Groq LLM API, based on the Llama~3 backbone. Prompt orchestration was handled via the \texttt{langchain\_groq} interface, defining separate templates for caption and hashtag generation. In every analyzed image, the YOLO detections, localized color descriptors, and retrieval-augmented features (fabric, gender) were serialised into a structured evidence pack and injected into the caption prompt. The instructing caption then asked the model to compose a 2-3 sentence, fluent, and image-driven caption that mentioned colors, fabrics, and garments that had been identified. The hashtag trigger then generated 15-18 varied tags of broad, mid-tier, and niche fashion keywords. The parameters of the model were adjusted to a temperature of approximately 0.7 and to a maximum length of output of approximately 250 tokens to ensure that there was creativity and coherence. In the absence of the LLM or API, a rule-based fallback generator generated descriptive sentences and simple hashtags based on identified classes, colors, and derived attributes to maintain continuous operation when using offline execution.

\subsection{BLIP Baseline Setup and Details}

The BLIP baseline was fine-tuned as a supervised vision–language model to provide a comparative reference for the proposed retrieval-augmented pipeline. This was implemented using the ``Salesforce/blip-image-captioning-base'' checkpoint from the Hugging~Face library, using the \texttt{AutoProcessor} and \texttt{BlipForConditionalGeneration} interfaces within the \texttt{transformers} framework. The same 80/20 category-balanced data split used for the RAG index was adopted to maintain consistency across experiments. Training was performed for five epochs using the AdamW optimizer with a learning rate $5\times10^{-5}$ and mixed-precision computation on an NVIDIA~A100 GPU. Each batch included paired image–caption samples processed into input token IDs and pixel embeddings. The model was optimized via cross-entropy loss between generated and reference captions, updating both visual and textual parameters end-to-end. The loss curves were tracked at both batch and epoch levels to verify the stability of the convergence, and the checkpoint with the lowest validation loss had been selected. In the inference phase, the fine-tuned BLIP model produced captions without retrieval or attribute conditioning using images as direct inputs.

The entire implementation comprises all modules in end-to-end pipeline. YOLO detection, color extraction, CLIP retrieval and LLM generation are sequentially run with unified settings, whereas the BLIP baseline is a supervised baseline. All stages were applied on modular scripts with fixed parameters and constant dataset splits making them comparable. This unified arrangement offers a firm basis on which the accuracy of perception and captioning performance could be assessed in the latter results section.

\section{Results}\label{sec:results}

\subsection{YOLO Detection Performance}

\begin{figure}[!t]
\centering
\includegraphics[width=0.48\textwidth]{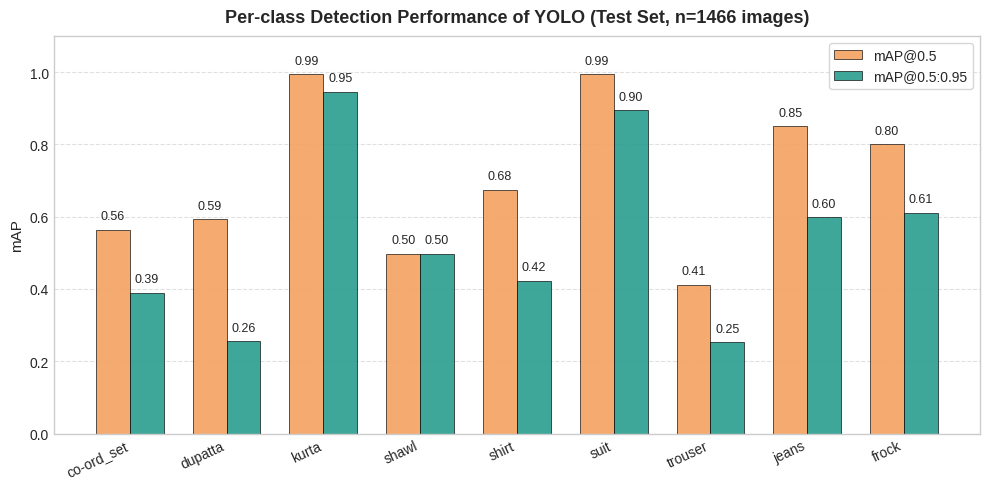}
\caption{Per-class detection performance of YOLO showing mAP@0.5 and mAP@0.5:0.95 across garment categories.}
\label{fig:yolo_map}
\end{figure}

The YOLO-based garment detector was evaluated on 1,466 test images containing 1,974 annotated instances across nine apparel categories. Figure~\ref{fig:yolo_map} presents the per-class detection performance in terms of mAP@0.5 and mAP@0.5:0.95. The overall mAP@0.5 reached 0.709, confirming reliable localization and classification of multiple garments within complex social and catalog scenes. Among individual categories, \textit{jeans} and \textit{kurta} achieved the highest detection accuracy with mAP@0.5 scores of 0.85 and 0.99 respectively, followed by \textit{frock} (0.80) and \textit{shirt} (0.67). Relatively lower scores for \textit{dupatta} and \textit{trouser} were attributed to fine-grained boundaries and occlusion, which occasionally caused partial detections or misclassifications. The corresponding mAP@0.5:0.95 values show a consistent decline across categories, reflecting the expected drop in performance under stricter localization thresholds.

\begin{figure}[!t]
\centering
\includegraphics[width=0.48\textwidth]{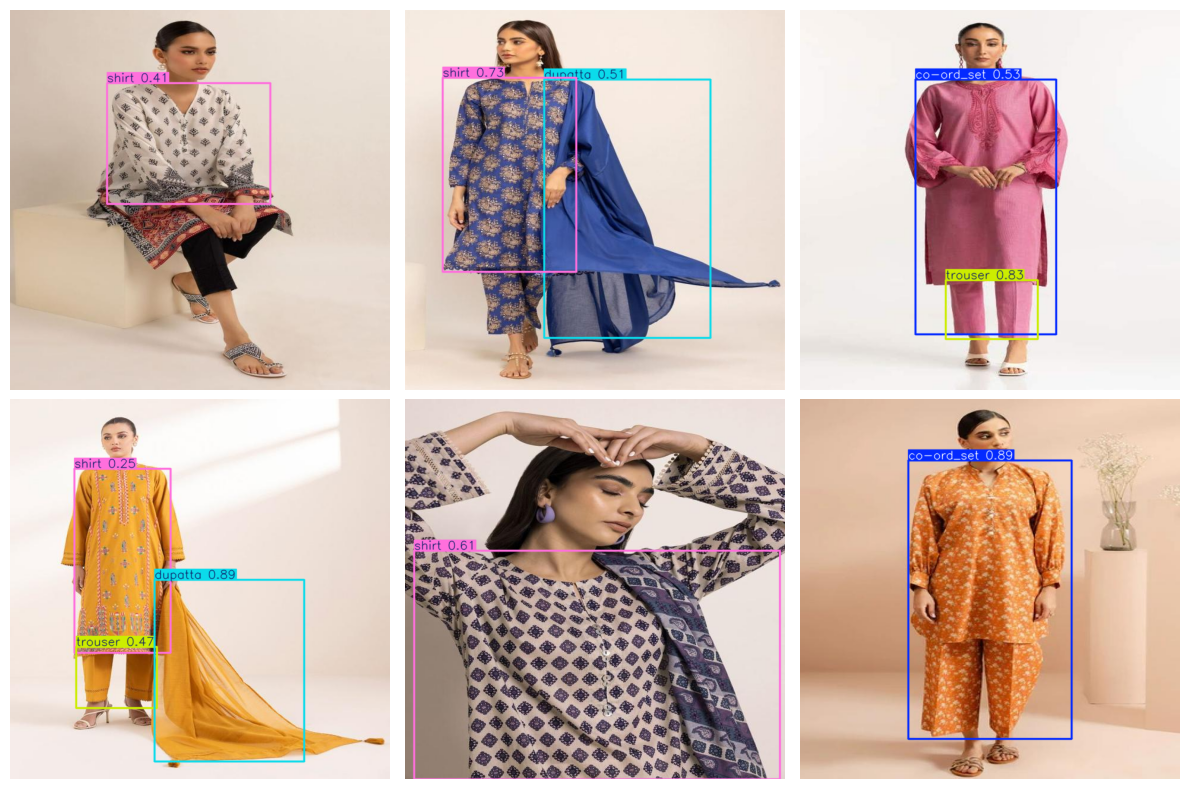}
\caption{Example detection outputs illustrating accurate multi-garment recognition in social and catalog scenes.}
\label{fig:yolo_samples}
\end{figure}

Figure~\ref{fig:yolo_samples} illustrates representative qualitative results demonstrating the model’s ability to detect multiple garments in a single frame. The detector successfully identifies overlapping apparel such as shirts, dupattas, and trousers while preserving their spatial arrangement. The results confirm that the fine-tuned YOLOv11s model generalizes effectively across lighting variations, pose diversity, and complex social backgrounds, providing a robust foundation for the subsequent captioning and retrieval modules.

\subsection{Caption Quality: BLIP vs RAG-LLM}

\begin{figure}[!t]
\centering
\includegraphics[width=0.48\textwidth]{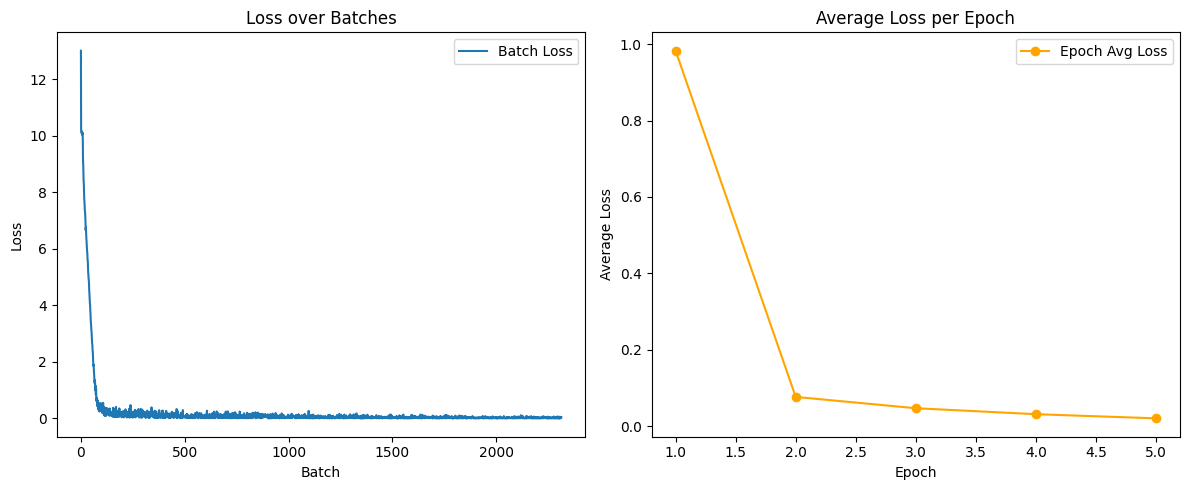}
\caption{BLIP fine-tuning loss curves showing batch-level and epoch-level training loss over five epochs.}
\label{fig:blip_loss}
\end{figure}

Caption generation results were evaluated for both the BLIP baseline and the proposed retrieval-augmented pipeline using BLEU, METEOR, ROUGE, and CLIP similarity metrics. Table~\ref{tab:caption_metrics} summarizes the quantitative performance across models. The fine-tuning process for BLIP converged smoothly, as illustrated in Fig.~\ref{fig:blip_loss}, where both batch-level and epoch-level losses steadily decreased across five epochs, indicating stable optimization without overfitting. The fine-tuned BLIP model achieved higher BLEU (0.2120), METEOR (0.5845), and ROUGE-L (0.4194) scores, indicating strong lexical alignment with the reference descriptions. These results reflect BLIP’s supervised learning behavior—memorizing the linguistic patterns present in the training captions—leading to syntactically accurate but templated outputs with limited stylistic variation. Because BLIP is fine-tuned for a specific dataset, expanding it to new clothing categories would require additional labeled data and retraining to maintain quality, limiting its scalability across broader fashion domains.

\begin{figure}[htbp]
\centering
\includegraphics[width=0.45\textwidth]{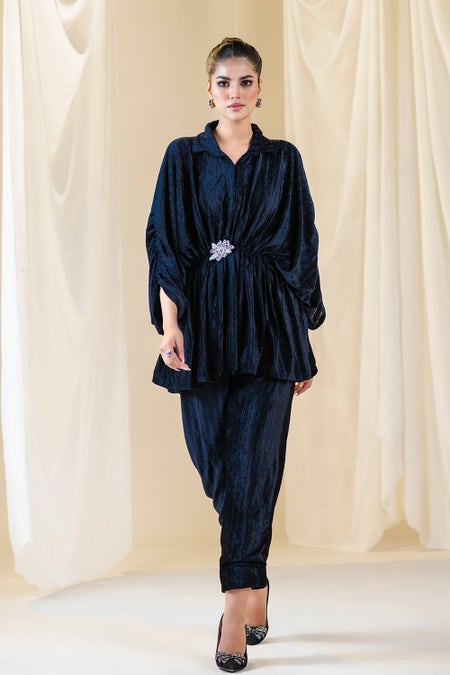}

\footnotesize
\textbf{Original Caption:} ``Chic and sophisticated, our Velvet Dyed Co-Ord Set in a deep blue hue features a velvet shirt paired with straight pants. This solid suit is perfect for making a stylish statement at any event.''\\[3pt]
\textbf{BLIP Generated Caption:} ``multicolorechic and sophisticated, our velvet dyed co - ord set in a deep blue hue features a velvet shirt paired with straight pants. this elegant suit is perfect for making a stylish statement at any event.''\\[3pt]
\textbf{RAG-LLM Generated Caption:} ``Elevate your evening style with our velvet co-ord set, featuring a black velvet shirt and straight pants that exude sophistication and comfort, perfect for a formal winter event.''\\[3pt]
\textbf{RAG-LLM Generated Hashtags:} ``\#FashionForWomen \#VelvetClothing \#WinterFashion \#FormalWear \#BlackOutfit \#CoOrdSet \#LuxuryFashion \#EveningStyle \#VelvetShirt \#StraightPants \#FemaleFashion \#SophisticatedStyle \#WinterFormalWear \#VelvetFashionTrends \#BlackVelvetOutfit \#Women'' \\[3pt]
\caption{Qualitative caption comparison for a sample test image showing the original brand description, BLIP baseline output, and RAG-LLM generated caption and hashtags.}
\label{fig:caption_qualitative}
\end{figure}

In contrast, the RAG-LLM pipeline scored lower on n-gram-based scores (BLEU~=~0.0230, METEOR~=~0.1374, ROUGE-L~=~0.1340) because its captions are not optimized around word-level overlap. It rather composes accounts based on recovered evidence of attributes and on stylistic exemplification, and produces fluent, human-like, and context-sensitive narratives based on the factual garment characteristics like color, fabric, and category. This formulation, which is founded on retrieval, enables the pipeline to extrapolate on missing or low-resource categories of fashions without further fine-tuning. The qualitative comparison in Fig.~\ref{fig:caption_qualitative} shows that the qualitative captions of RAG-LLM are more contextualized and descriptively realistic than those of BLIP.

The CLIP similarity analysis provides complementary insight into semantic grounding. For BLIP, the mean similarity between image and predicted caption (0.3134) slightly exceeded that of the original product descriptions (0.3098), indicating close visual–semantic alignment but minimal linguistic novelty. RAG-LLM captions, while showing a lower similarity (0.2827) relative to the originals (0.3102), maintain factual grounding while achieving greater linguistic diversity. This trade-off is desirable for creative fashion narratives, where human-style expression is prioritized over exact textual replication. Overall, BLIP excels in lexical fidelity for known classes, whereas RAG-LLM delivers more generalizable, attribute-driven, and hallucination-resistant captions suitable for large-scale automated fashion content generation.

\begin{table*}[!t]
\centering
\caption{Caption quality comparison between BLIP and RAG-LLM using BLEU, METEOR, ROUGE (F-scores), and CLIP similarity.}
\label{tab:caption_metrics}
\setlength{\tabcolsep}{6pt}
\renewcommand{\arraystretch}{1.05}
\footnotesize
\begin{tabular}{lccccccc}
\hline
\textbf{Model} & \textbf{BLEU} & \textbf{METEOR} & \textbf{R1-F} & \textbf{R2-F} & \textbf{RL-F} & \textbf{CLIP$_{\text{pred}}$} \\
\hline
BLIP     & 0.2120 & 0.5845 & 0.4336 & 0.3058 & 0.4194 & 0.3134 \\
RAG-LLM  & 0.0230 & 0.1374 & 0.1556 & 0.0358 & 0.1340 & 0.2827 \\
\hline
\end{tabular}
\end{table*}

\begin{figure}[!t]
\centering
\includegraphics[width=0.48\textwidth]{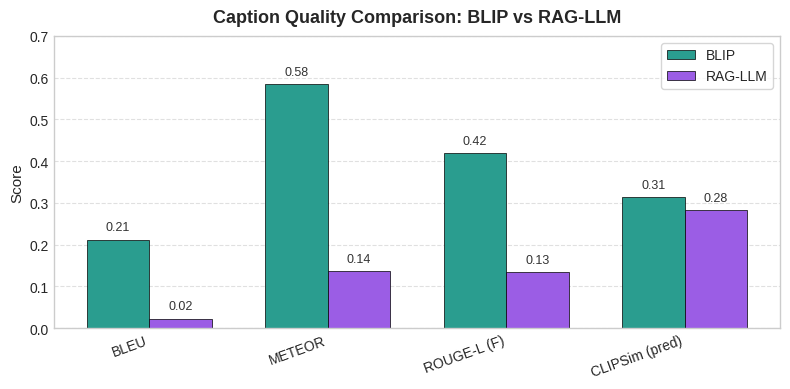}
\caption{Comparison of caption quality metrics (BLEU, METEOR, ROUGE-L, CLIP similarity) for BLIP and RAG-LLM.}
\label{fig:caption_metrics_bar}
\end{figure}

\subsection{Hashtag Evaluation}

The generated hashtags from the RAG-LLM pipeline were evaluated using the attribute coverage and diversity metrics described in Section~3.6. Quantitative results are summarized in Table~\ref{tab:hashtag_metrics}. The system achieved exceptionally high coverage across all evaluated thresholds, with a mean attribute coverage of 0.7976 and Coverage@0.5 equal to 1.000, indicating that every test image contained hashtags reflecting at least half of its key visual facets (garment category, dominant color, fabric, and gender). Even under stricter thresholds, Coverage@0.6 and Coverage@0.7 remained above 0.98, confirming that nearly all outputs correctly captured 60–70\% of their corresponding attributes. These results highlight the factual grounding and attribute consistency of the retrieval-augmented generation framework, where the structured evidence pack ensures that generated hashtags accurately represent the visual content.

Distinct-$n$ ratios were used to measure the linguistic diversity of the produced hashtags. The score of 0.0401 in Distinct-1 and 0.3188 in Distinct-2 indicate a moderate level of the lexical variation at the unigram level but high bigram diversity, indicating that individual tokens (e.g., fashion, style) are repeated across pictures but their combinations create context-rich and unique expressions. Such a high attribute coverage and moderate linguistic diversity show that the RAG-LLM paradigm produces hashtags that are semantically faithful but stylistically diverse and can be applied to the real-life fashion marketing and social media use.

\begin{table*}[!t]
\centering
\caption{Quantitative results for hashtag generation showing attribute coverage and diversity metrics.}
\label{tab:hashtag_metrics}
\setlength{\tabcolsep}{8pt}
\renewcommand{\arraystretch}{1.05}
\footnotesize
\begin{tabular}{lcccccc}
\hline
\textbf{Total Images} & \textbf{Mean Coverage} & \textbf{Coverage@0.5} & \textbf{Coverage@0.6} & \textbf{Coverage@0.7} & \textbf{Distinct-1} & \textbf{Distinct-2} \\
\hline
226 & 0.7976 & 1.000 & 0.9823 & 0.9823 & 0.0401 & 0.3188 \\
\hline
\end{tabular}
\end{table*}

\section{Discussion and Future Work}\label{sec:D&F}

The overall results of the experiment prove that the suggested retrieval-augmented captioning pipeline is effective in balancing between structured attribute reasoning and human-like narrative generation. All of these components, object detection, attribute retrieval, and language generation, are involved in creating visually faithful and stylistically fluent results that can be used in the real world in fashion segment. The YOLO-based detector was shown to be highly accurate among various garment classes and also capable of detecting various apparel objects with high reliability even when used in social media images with a complex background. High detection consistency was a direct support of downstream captioning and hashtag steps, whereby each crop and its properties gave factual support to text generation.

The BLIP baseline versus the RAG-LLM pipeline comparison shows that there is a definite trade-off between the lexical fidelity and the contextual expressiveness. Trained in a supervised way on paired brand descriptions, BLIP scored high on BLEU, METEOR, and ROUGE, which means that it is related closely to the reference corpus. Nevertheless, such captions are likely to mimic the syntactic patterns of the fine-tuning process, which leads to the repetitive syntactic patterns and the lack of stylistic diversity. Since BLIP relies on a fixed dataset, it would need to introduce more labeled text-image pairs and retrain to be applied to new clothing categories or new regions of fashion, which limits its flexibility.

Conversely, the retrieval-augmented approach extrapolates beyond the particular categories observed in the construction of the index. The RAG-LLM system builds descriptions based on visual detection (category, color) and retrieval-based inference (fabric, gender) attributes, which make them factually accurate on previously unseen garments. The qualitative results indicate that RAG-LLM captions are more likely to be more marketing-like in style with a focus on texture, coordination, and aesthetic context instead of direct reproduction of catalog language. Even though its n-gram overlap with reference text is less, captions are characterized by high CLIP similarity, which proves that visual-semantic coherence is maintained. Such behavior is indicative of the fact that retrieval conditioning is useful in reducing hallucination and improving factual consistency, which is a desired attribute of fashion e-commerce and content creation pipelines.

The hashtag evaluation further validates these strengths. The proposed attribute-coverage metric shows that nearly all generated hashtags represent at least 60–70\% of the visual facets, while the moderate Distinct-$n$ scores confirm balanced diversity without uncontrolled randomness. These results highlight that the model not only captures the key product attributes but also presents them through varied and socially relevant language patterns, supporting engagement-oriented applications such as automated post generation or trend analysis.

Even though it has its benefits, the proposed pipeline has some drawbacks. The quality of the indexed catalog and its representativeness are very important in the retrieval stage; noise and biased metadata may be transferred to the generated captions. On the same note, the detector can be generalized well in the domain of South Asian fashion, but it might need retraining or inclusion of additional classes to address other types of garments or accessories. The existing CLIP and LLM modules can also be used separately; more intimate multimodal interaction would provide additional improvements to visual-text alignment. In addition, the current system only accepts English captions, which restricts its use in multilingual fashion markets.

The future work will be oriented to three main directions. To begin with, the retrieval index should be expanded with bigger and more varied brand sets to enhance the robustness of attributes and their stylistic diversity. Second, a wider regional adoption could be made possible by incorporating multilingual caption generation through cross-lingual retrieval and LLM prompting. Third, the CLIP encoder and the captioning model can be fine-tuned together in a single retrieval-generation loop, which could enhance semantic cohesion and decrease reliance on fixed metadata. Lastly, user engagement metrics (like click-through or like rates) would offer a viable authentication of caption and hashtag efficiency in social settings.

\section{Conclusion}\label{sec:conclusion}

This work introduced a retrieval-augmented system of fashion caption and hashtag generation that combines multiple garment detection, visual attribute justification, and LLM prompting. The system exhibited high detection, factual grounding, and stylistic fluency in multifaceted fashion pictures. The RAG-LLM pipeline generated more human-like, attribute-centric, and generalizable captions than the fine-tuned BLIP baseline and was semantically aligned with visual data. The reliability of generated hashtags was also confirmed by the proposed attribute coverage and diversity measure. Collectively, these findings demonstrate the opportunity of retrieval-enhanced generation as a scalable remedy to visually grounded, socially versatile content creation in fashion. Further extensions to multilingual captioning, larger retrieval corpora and unified multimodal optimization can improve its performance and practical use in real-world implementation.

\section*{Data Availability} The data used for this research will be made available upon request.

\section*{Funding} This research did not receive any specific grant from funding agencies in the public, commercial, or not-for-profit sectors.

\bibliographystyle{elsarticle-num}   
\bibliography{cas-refs}              

@ARTICLE{thakare2024survey,
  author  = {Thakare, Yugandhara A. and Walse, Kishor H.},
  title   = {A review of Deep learning image captioning approaches},
  journal = {Journal of Integrated Science \& Technology},
  volume  = {12},
  number  = {1},
  year    = {2024},
  pages   = {712},
}

@ARTICLE{xu2023deep,
  author  = {Xu, Liming and Tang, Quan and Lv, Jiancheng and Zheng, Bochuan and Zeng, Xianhua and Li, Weisheng},
  title   = {Deep image captioning: A review of methods, trends and future challenges},
  journal = {Neurocomputing},
  volume  = {546},
  year    = {2023},
  pages   = {126287}
}

@ARTICLE{jia2023scenegraphs,
  author  = {Jia, Junhua and Ding, Xiangqian and Pang, Shunpeng and Gao, Xiaoyan and Xin, Xiaowei and Hu, Ruotong and Nie, Jie},
  title   = {Image captioning based on scene graphs: A survey},
  journal = {Expert Systems with Applications},
  volume  = {231},
  year    = {2023},
  pages   = {120698}
}

@CONFERENCE{gusain2023instacaption,
  author    = {Gusain, Ravi and Pathak, Saksham and Ghosh, Soumi},
  title     = {Instagram Post Caption Recommendation Engine},
  booktitle = {Proceedings of the 14th International Conference on Computing Communication and Networking Technologies (ICCCNT)},
  pages     = {1--3},
  publisher = {IEEE},
  year      = {2023}
}

@ARTICLE{jafarisadr2023tags,
  author  = {Jafari Sadr, Mohammad and Mirtaheri, Seyedeh Leili and Greco, Sergio and Borna, Keivan},
  title   = {Popular Tag Recommendation by Neural Network in Social Media},
  journal = {Computational Intelligence and Neuroscience},
  volume  = {2023},
  number  = {1},
  pages   = {4300408},
  publisher = {Wiley Online Library},
  year    = {2023}
}

@ARTICLE{balloni2024social4fashion,
  author  = {Balloni, Emanuele and Pietrini, Rocco and Fabiani, Matteo and Frontoni, Emanuele and Mancini, Adriano and Paolanti, Marina},
  title   = {Social4Fashion: An intelligent expert system for forecasting fashion trends from social media contents},
  journal = {Expert Systems with Applications},
  volume  = {252},
  pages   = {124018},
  publisher = {Elsevier},
  year    = {2024}
}

@ARTICLE{islam2024fashionretrieval,
  author  = {Islam, Sk Maidul and Joardar, Subhankar and Sekh, Arif Ahmed},
  title   = {A Survey on Fashion Image Retrieval},
  journal = {ACM Computing Surveys},
  volume  = {56},
  number  = {6},
  pages   = {1--25},
  publisher = {ACM},
  address = {New York, NY},
  year    = {2024}
}

@ARTICLE{tang2023fashionattr,
  author  = {Tang, Yuhao and Zhang, Liyan and Yuan, Ye and Chen, Zhixian},
  title   = {Improving Fashion Captioning via Attribute-Based Alignment and Multi-Level Language Model},
  journal = {Applied Intelligence},
  volume  = {53},
  number  = {24},
  pages   = {30803--30821},
  publisher = {Springer},
  year    = {2023}
}

@ARTICLE{lavanya2024yolo,
  author  = {Lavanya, Gudala and Pande, Sagar Dhanraj},
  title   = {Enhancing Real-time Object Detection with YOLO Algorithm},
  journal = {EAI Endorsed Transactions on Internet of Things},
  volume  = {10},
  year    = {2023}
}

@ARTICLE{kang2023yolo,
  author  = {Kang, Chang Ho and Kim, Sun Young},
  title   = {Real-time Object Detection and Segmentation Technology: An Analysis of the YOLO Algorithm},
  journal = {JMST Advances},
  volume  = {5},
  number  = {2},
  pages   = {69--76},
  publisher = {Springer},
  year    = {2023}
}

@ARTICLE{he2023fashionmaskrcnn,
  author  = {He, Wentao and Wang, Jing'an and Wang, Lei and Pan, Ruru and Gao, Weidong},
  title   = {A Semantic Segmentation Algorithm for Fashion Images Based on Modified Mask RCNN},
  journal = {Multimedia Tools and Applications},
  volume  = {82},
  number  = {18},
  pages   = {28427--28444},
  publisher = {Springer},
  year    = {2023}
}

@ARTICLE{chang2022styleyolo,
  author  = {Chang, Yeong-Hwa and Zhang, Ya-Ying},
  title   = {Deep Learning for Clothing Style Recognition Using YOLOv5},
  journal = {Micromachines},
  volume  = {13},
  number  = {10},
  pages   = {1678},
  publisher = {MDPI},
  year    = {2022}
}

@ARTICLE{sirisha2022acrossdomains,
  author  = {Sirisha, Uddagiri and Chandana, Bolem Sai},
  title   = {Semantic Interdisciplinary Evaluation of Image Captioning Models},
  journal = {Cogent Engineering},
  volume  = {9},
  number  = {1},
  pages   = {2104333},
  publisher = {Taylor \& Francis},
  year    = {2022}
}

@ARTICLE{khurram2021densecaps,
  author  = {Khurram, Imran and Fraz, Muhammad Moazam and Shahzad, Muhammad and Rajpoot, Nasir M.},
  title   = {Dense-CaptionNet: A Sentence Generation Architecture for Fine-Grained Description of Image Semantics},
  journal = {Cognitive Computation},
  volume  = {13},
  pages   = {595--611},
  publisher = {Springer},
  year    = {2021}
}

@ARTICLE{rinaldi2023multinetwork,
  author  = {Rinaldi, Antonio M. and Russo, Cristiano and Tommasino, Cristian},
  title   = {Automatic Image Captioning Combining Natural Language Processing and Deep Neural Networks},
  journal = {Results in Engineering},
  volume  = {18},
  pages   = {101107},
  publisher = {Elsevier},
  year    = {2023}
}

@ARTICLE{ghandi2023deep,
  author  = {Ghandi, Taraneh and Pourreza, Hamidreza and Mahyar, Hamidreza},
  title   = {Deep Learning Approaches on Image Captioning: A Review},
  journal = {ACM Computing Surveys},
  volume  = {56},
  number  = {3},
  publisher = {ACM},
  address = {New York, NY, USA},
  year    = {2023}
}

@ARTICLE{nursikuwagus2022geocap,
  author  = {Nursikuwagus, Agus and Munir, Rinaldi and Khodra, Masayu Leylia},
  title   = {Hybrid of Deep Learning and Word Embedding in Generating Captions: Image-Captioning Solution for Geological Rock Images},
  journal = {Journal of Imaging},
  volume  = {8},
  number  = {11},
  pages   = {294},
  publisher = {MDPI},
  year    = {2022}
}

@ARTICLE{li2024bridgecaption,
  author  = {Li, Shunlong and Dang, Minghao and Xu, Yang and Wang, Andong and Guo, Yapeng},
  title   = {Bridge Damage Description Using Adaptive Attention-Based Image Captioning},
  journal = {Automation in Construction},
  volume  = {165},
  pages   = {105525},
  publisher = {Elsevier},
  year    = {2024}
}

@ARTICLE{han2024vitvgg,
  author  = {Han, Huimin and Aboubakar, Bouba Oumarou and Bhatti, Mughair and Talpur, Bandeh Ali and Ali, Yasser A. and Al-Razgan, Muna and Ghadi, Yazeed Yasid},
  title   = {Optimizing Image Captioning: The Effectiveness of Vision Transformers and VGG Networks for Remote Sensing},
  journal = {Big Data Research},
  volume  = {37},
  pages   = {100477},
  publisher = {Elsevier},
  year    = {2024}
}

@CONFERENCE{elsady2023remotesensing,
  author    = {Elsady, Reem Mostafa and Ahmed, Youssef Abdelrahman and Salem, Mohammed Abdel-Megeed},
  title     = {Remote Sensing Image Segmentation and Captioning Using Deep Learning},
  booktitle = {Proceedings of the 2nd International Conference on Smart Cities 4.0},
  pages     = {196--201},
  publisher = {IEEE},
  year      = {2023}
}

@ARTICLE{he2022segmentation,
  author  = {He, Defu and Xie, Chao},
  title   = {Semantic Image Segmentation Algorithm in a Deep Learning Computer Network},
  journal = {Multimedia Systems},
  volume  = {28},
  number  = {6},
  pages   = {2065--2077},
  publisher = {Springer-Verlag},
  year    = {2022}
}

@ARTICLE{tereikovskyi2022method,
  author  = {Tereikovskyi, Ihor and Hu, Zhengbing and Chernyshev, Denys and Tereikovska, Liudmyla and Korystin, Oleksandr and Tereikovskyi, Oleh},
  title   = {The Method of Semantic Image Segmentation Using Neural Networks},
  journal = {International Journal of Image, Graphics and Signal Processing},
  volume  = {10},
  number  = {6},
  pages   = {1},
  publisher = {Modern Education and Computer Science Press},
  year    = {2022}
}

@ARTICLE{wang2022construction,
  author  = {Wang, Yiheng and Xiao, Bo and Bouferguene, Ahmed and Al-Hussein, Mohamed and Li, Heng},
  title   = {Vision-Based Method for Semantic Information Extraction in Construction by Integrating Deep Learning Object Detection and Image Captioning},
  journal = {Advanced Engineering Informatics},
  volume  = {53},
  pages   = {101699},
  publisher = {Elsevier},
  year    = {2022}
}

@ARTICLE{mahalakshmi2022summarization,
  author  = {Mahalakshmi, P. and Fatima, N. Sabiyath},
  title   = {Summarization of Text and Image Captioning in Information Retrieval Using Deep Learning Techniques},
  journal = {IEEE Access},
  volume  = {10},
  pages   = {18289--18297},
  publisher = {IEEE},
  year    = {2022}
}

@ARTICLE{choi2024irsn,
  author  = {Choi, Jinyoung and Kwon, Youngchae and Kim, Injung},
  title   = {Item-Region-Based Style Classification Network (IRSN): A Fashion Style Classifier Based on Domain Knowledge of Fashion Experts},
  journal = {Applied Intelligence},
  volume  = {54},
  number  = {20},
  pages   = {9579--9593},
  publisher = {Kluwer Academic Publishers},
  year    = {2024}
}

@ARTICLE{kalinin2024stylegen,
  author  = {Kalinin, Aleksandr and Jafari, Akbar Anbar and Avots, Egils and Ozcinar, Cagri and Anbarjafari, Gholamreza},
  title   = {Generative AI-Based Style Recommendation Using Fashion Item Detection and Classification},
  journal = {Signal, Image and Video Processing},
  pages   = {9179--9189},
  volume = {18},
  publisher = {Springer},
  year    = {2024}
}

@ARTICLE{xu2023crossmodal,
  author  = {Xu, Chunpu and Tan, Hanzhuo and Li, Jing and Li, Piji},
  title   = {Understanding Social Media Cross-Modality Discourse in Linguistic Space},
  journal = {arXiv preprint arXiv:2302.13311},
  year    = {2023}
}

@ARTICLE{alduhayyim2022metaic,
  author  = {Al Duhayyim, Mesfer and Alazwari, Sana and Mengash, Hanan Abdullah and Marzouk, Radwa and Alzahrani, Jaber S. and Mahgoub, Hany and Althukair, Fahd and Salama, Ahmed S.},
  title   = {Metaheuristics Optimization with Deep Learning Enabled Automated Image Captioning System},
  journal = {Applied Sciences},
  volume  = {12},
  number  = {15},
  pages   = {7724},
  publisher = {MDPI},
  year    = {2022}
}

@CONFERENCE{yu2022metaic,
  author    = {Yu, Boxi and Zhong, Zhiqing and Qin, Xinran and Yao, Jiayi and Wang, Yuancheng and He, Pinjia},
  title     = {Automated Testing of Image Captioning Systems},
  booktitle = {Proceedings of the 31st ACM SIGSOFT International Symposium on Software Testing and Analysis},
  pages     = {467--479},
  year      = {2022}
}

\end{document}